\documentclass[lettersize,journal]{IEEEtran}
\usepackage{cite}
\usepackage{amsmath,amssymb,amsfonts}
\usepackage{algorithm}
\usepackage{algorithmic}
\usepackage{graphicx}
\usepackage{float}
\usepackage{textcomp}
\usepackage{xcolor}
\usepackage{subfigure}
\usepackage{caption}
\usepackage{hyperref}
\usepackage{setspace}
\usepackage{tablefootnote}

\def\BibTeX{{\rm B\kern-.05em{\sc i\kern-.025em b}\kern-.08em
    T\kern-.1667em\lower.7ex\hbox{E}\kern-.125emX}}
\begin{document}

\title{A Hybrid Framework of Reinforcement Learning and Convex Optimization for UAV-Based Autonomous Metaverse Data Collection}
\author{\IEEEauthorblockN{Peiyuan Si,
Liangxin Qian,
Jun Zhao,
Kwok-Yan Lam\\}
\IEEEauthorblockA{School of Computer Science \& Engineering, Nanyang Technological University, Singapore\\
\{peiyuan001, qian0080\}@e.ntu.edu.sg, \{junzhao,  kwokyan.lam\}@ntu.edu.sg }}

\maketitle
\markboth{This paper appears in IEEE Network magazine.}{}

\begin{abstract}
Unmanned aerial vehicles (UAVs) are promising for providing communication services due to their advantages in cost and mobility, especially in the context of the emerging Metaverse and Internet of Things (IoT). This paper considers a UAV-assisted Metaverse network, in which UAVs extend the coverage of the base station (BS) to collect the Metaverse data generated at roadside units (RSUs). Specifically, to improve the data collection efficiency, resource allocation and trajectory control are integrated into the system model. The time-dependent nature of the optimization problem makes it non-trivial to be solved by traditional convex optimization methods. Based on the proposed UAV-assisted Metaverse network system model, we design a hybrid framework with reinforcement learning and convex optimization to {cooperatively} solve the time-sequential optimization problem.
Simulation results show that the proposed framework is able to reduce the mission completion time with a given transmission power resource.
\end{abstract}

\section{Introduction}
% Background of Metaverse: include RSU & IoT (2)
Metaverse is an emerging concept that aims to generate the digital duality of the physical world and provide users with immersive experience through augmented reality (AR), virtual reality (VR), and other novel techniques \cite{cheng2022will}.
To ensure the quality of service (QoS), 5G or future 6G communication technologies are needed to support the required transmission rate of AR/VR service. However, the improvement of transmission rate is at the cost of the coverage area of one single base station (BS) due to severe attenuation of signals, which makes it more expensive to provide wireless communication coverage. On the other hand, the maintenance and update of the virtual world in Metaverse is based on massive data collected by individual devices, roadside units (RSUs), and other Internet of Things (IoT) devices. Thus, it is prominent to extend the coverage of Metaverse data collection and synchronization to provide more users with seamless service.

% connected vehicle network
{Connected vehicle network is a promising solution to extend the coverage of Metaverse data collection by utilizing the sensors and computation capacities of vehicles.
In traditional connected vehicle networks, roadside units (RSUs) are fundamental devices for data storage. Reliable and efficient data transfer from RSUs to the server is required for Metaverse data collection, but the coverage of high speed wireless communication can be costly in the suburb with low population and data density.}
%UAV data collection in Metaverse (5)
To collect the data stored at RSUs, unmanned aerial vehicles (UAVs) assisted communication has attracted attention to obtaining cheap and flexible Metaverse data collection \cite{Vision_UAV_meta, uav_base_station}. The mobility and load-carrying ability of UAV makes it a distinguished platform for all kinds of devices to assist in Metaverse service, such as data synchronization, sensing, and mobile base station services \cite{uav_data_collect_survey,metaverse_uav}.
{With the support of UAVs, efficient autonomous data collection from connected vehicles and RSUs can be implemented to extend the coverage of Metaverse, as UAVs can reduce the distance to the data transmitter for better channel condition and alleviate signal attenuation.} Besides, UAV-assisted wireless power transfer (WPT) can be conducted to charge the sensors deployed at roadside units (RSUs) wirelessly, and thus prolong the sensor network lifetime.

% Discussion about UAV resource allocation optimization problems (2)
The existing research on UAV-assisted data collection mainly focuses on the resource allocation and UAV trajectory planning for two objectives: 1) reducing energy consumption \cite{energy_aware_UAV}; 2) reducing the required time for data collection \cite{time_limit_UAV}.
The specific optimization problem can be diverse under different system models and objectives, but there are some common issues and challenges in resource allocation and trajectory optimization.
Firstly, trajectory optimization introduces the distance between UAVs and ground devices as a variable, which makes the problem non-convex due to the characteristic of the log function.
Secondly, the system models involving channel allocation inevitably introduce integer variables into the optimization problem and generates mixed-integer non-linear programming problems (MINLP). Although there are some methods to solve MINLP problems, such as the outer approximation algorithm and branch and bound algorithm, most existing algorithms are mainly based on the assumption of separable functions \cite{MINLP_survey}. The objective functions and constraints in the context of resource allocation and trajectory planning easily becomes inseparable and non-convex, which makes it non-trivial to obtain the optimal solution.
In addition to the difficulty in solving non-convex MINLP problems with inseparable functions, the flight time is usually divided into multiple time slots for computational convenience, which results in the time-sequential nature of trajectory planning problems. To obtain the best performance, the variables need to be optimized respectively in each time slot, i.e., an independent set of variables need to be defined for each time slot, which increases the complexity of algorithms and the difficulty of convergence.

% Application of RL in optimization (4)
Reinforcement learning (RL) aims to achieve the highest long-term reward in a time sequential system, which is a compatible alternative solution to UAV trajectory and communication resource allocation problems. In a classical RL system model, the agent interacts with the environment to receive a corresponding reward in each step. The objective of the RL agent is to maximize the sum reward in the whole episode which consists of multiple steps. This is similar to the UAV trajectory and resource allocation optimization problems, where the whole flight time is divided into multiple time slots, and the controller makes decisions in each time slot. Motivated by the similarity in the aim of RL and time-sequential communication problems, the application of RL in resource allocation and trajectory optimization have attracted much research interest, and the feasibility has been proved by the existing works \cite{rl_v2v, RL_multi}.

% Discussion about convex and RL (2)
The major advantages of RL over traditional convex optimization methods are: 1) It ensures a feasible solution with relatively good performance even if the theoretical optimal solution is hard to obtain; 2) The number of time slots does not influence the complexity of the RL agent design. However, RL can not ensure the best performance in some simple optimization problems whose global optimal can be found by convex optimization. Moreover, the exploration mechanisms of RL agents such as $\varepsilon$-greedy force it to make sub-optimal decisions with a small probability after achieving convergence, which degrades the stability of performance in the early period of the training process.

% Motivation: utilize the advantage of both
{Motivated by the strengths and weaknesses of RL and convex optimization methods, we propose a hybrid framework in which a convex solver and RL agents work cooperatively to solve the resource allocation and trajectory optimization problem for better performances in UAV-assisted Metaverse data collection.} The original non-convex MINLP optimization problem is decomposed into three sub-problems: channel allocation, power control, and trajectory optimization. Each sub-problem is solved by an RL agent or convex solver according to their characteristic. Simulation results show the superiority of the proposed hybrid RL and convex optimization framework with the proximal policy optimization algorithm over the benchmark algorithms.

\section{Hybrid Framework with Reinforcement Learning and Convex Optimization: A Case Study}

In this section, we give an example of designing hybrid frameworks with reinforcement learning and convex optimization for UAV-assisted data collection in Metaverse.
\subsection{System Architecture}

\begin{figure*}[htb]
 \centering
 \includegraphics[width=0.9\linewidth]{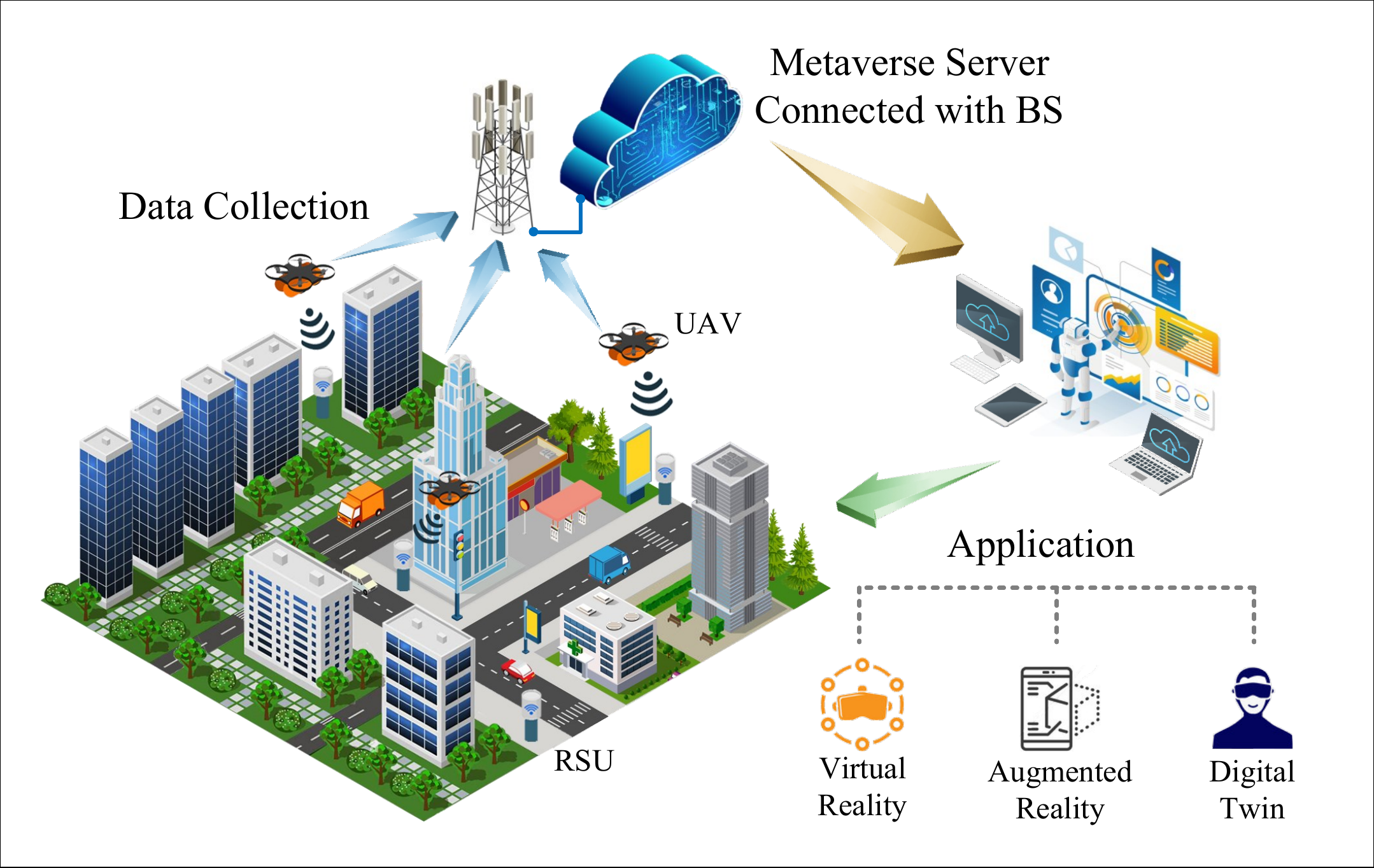}
 \caption{System architecture.}
 \label{fig:System_Model}
\end{figure*}

% brief on system model
The architecture of the UAV-assisted data collection model in the Metaverse network is shown in Fig. \ref{fig:System_Model}. Multiple roadside units (RSUs) with data storage are deployed in the urban area to gather environmental information such as traffic conditions, weather, and the change of building structures for the update of digital twin in the Metaverse. One or more UAVs are set out from the BS to collect the data cached at the RSUs beyond the coverage. Once the data collection mission is completed, the UAVs return to the BS and upload the data to the Metaverse server for data synchronization and services such as virtual reality (VR), augmented reality (AR), and digital twin.

% what we want to do
Due to the limitation of battery, the UAV needs to finish the data collection mission within a given flight time. Thus, it is important to improve the data collection efficiency, i.e., to reduce the mission completion time. In this paper, we aim to reduce the required time for data collection missions by optimizing the following variables: channel allocation, RSU transmission power, and UAV trajectory.
\begin{itemize}
\item
% channel allocation
\textbf{Channel allocation. }The wireless bandwidth resource is limited in practice, in this paper we assume that the RSUs can transmit data to UAVs over a limited number of channels, and the bandwidth of each channel is denoted as $B$.
Although there is no strict constraint on how many RSUs can transmit data over a single channel at the same time, it is not wise to allocate too many RSUs into the same channel because severe interference will degrade the data rate. Thus, there is a balance between the number of RSUs in the channel and the actual data rate, which is critical to reducing the mission time.
\item
% power
\textbf{Power control. }The RSUs are assumed to have a stable energy supply but limited transmission power. Increasing the transmission power of a specific RSU can improve its data rate, but the data rate of other RSUs in the same channel will be degraded due to severe interference, which could result in a decrease in the sum data rate. Thus, there is a balance between the transmission power of each RSU and the sum data rate.
\item
% trajectory
\textbf{Trajectory optimization. }The trajectory of the UAV is closely related to the distance between the sender (RSU) and receiver (UAV), which has a great impact on the channel gain. A smaller distance between the UAV to the RSU means less attenuation of the radio signal and a higher data rate, but in some cases, it is better to go through the middle of two RSUs for a higher total data rate. The UAV trajectory is also influenced by the channel gain and transmission power.
\end{itemize}

{In our implementation, the channel allocation should be decided before trajectory and power control because the calculation is based on given channel allocation. The power control is decided after the channel allocation, and finally is the trajectory control. The trajectory should be updated after the power control decision because we need to keep the location of UAV the same in the calculation in each iteration.}
In order to achieve the minimum required time for data collection missions, the hybrid framework with reinforcement learning and convex optimization for UAV-assisted data collection in Metaverse networks is proposed and introduced in the next section.

\subsection{Proposed Hybrid Framework for UAV-assisted Data Collection}

\begin{figure*}[htb]
 \centering
 \includegraphics[width=0.9\linewidth]{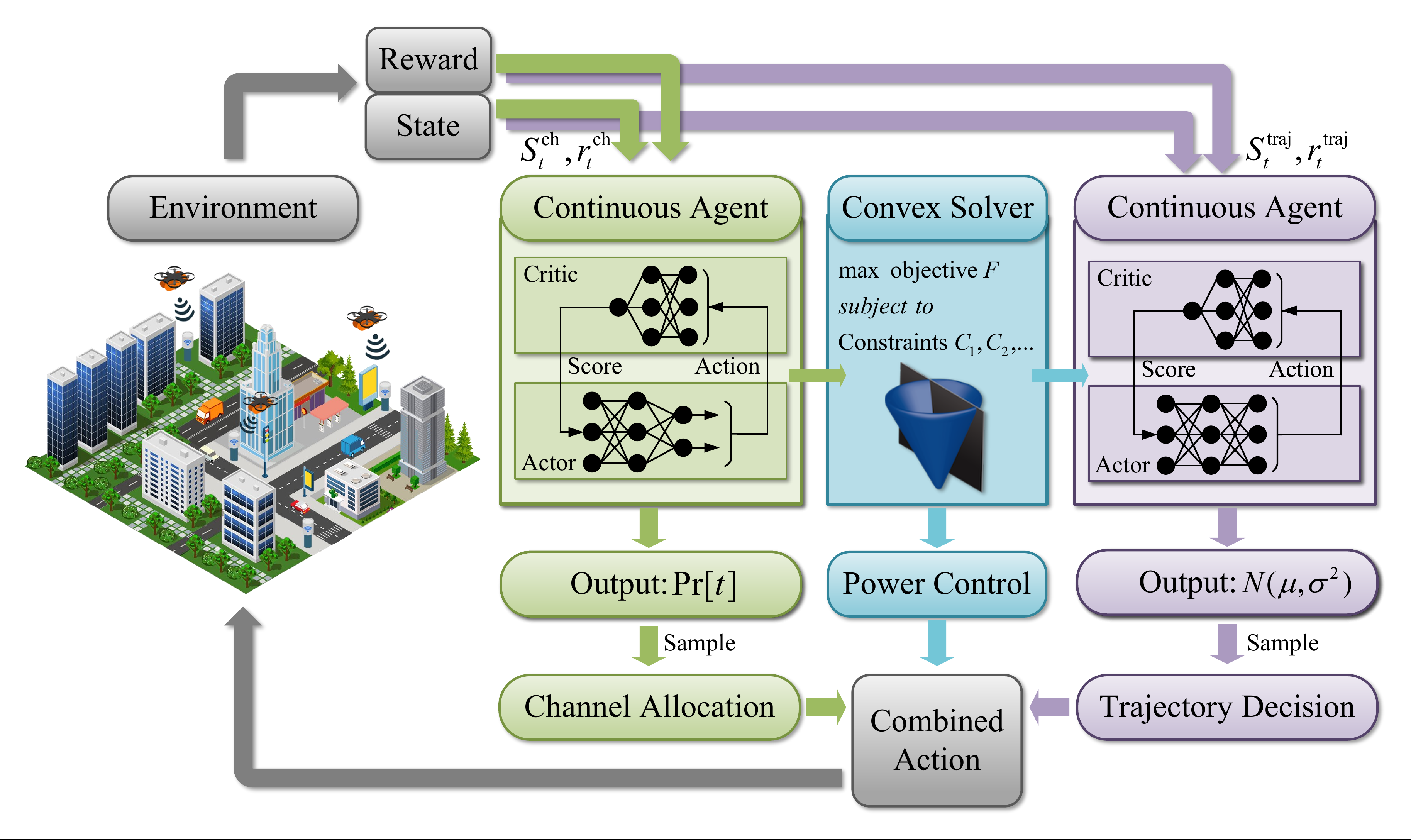}
 \caption{The proposed hybrid framework with reinforcement learning and convex optimization.}
 \label{fig:alg}
\end{figure*}
% general
The proposed hybrid framework consists of two reinforcement learning (RL) agents and a solver for convex optimization, which is shown in Fig. \ref{fig:alg}. An agent for trajectory optimization with continuous action space (continuous agent), an agent for channel allocation with discrete action space (discrete agent), and a convex solver for power control work {cooperatively} to generate the combined trajectory, power control, and channel allocation decision. The combined decision is adopted by the UAV and RSU to interact with the environment, which returns rewards to the RL agents for training. The RL agents learn from the feedback (reward) of the environment and gradually improve their policies over the training episodes, and finally converge at a policy with high performance.
The convex solver is able to obtain stable and optimal power control decisions as it always ensures the global optimal in each iteration.

% about a-c
The RL agents in the proposed framework adopt the widely-used actor-critic architecture, in which the ``actor" network makes actions according to the input state at $t^\text{th}$ time slot ($S_t^\text{Traj}$ for trajectory optimization agent and $S_t^\text{ch}$ for channel allocation agent) to get the highest evaluation from the ``critic" network. The ``critic" network aims to minimize the distance between its own evaluation and the ground truth reward given by the environment \cite{actor-critic, PPO}. In the training process, both the ``actor" and the ``critic" networks are updated by back propagation (BP). The difference is that the gradient to update the ``actor" network is generated by the evaluation (output) of the ``critic" network, while the gradient for the update of the ``critic" network is generated by the distance between its evaluation and the reward given by the environment.

In our proposed framework, there are some differences between the continuous agent and the discrete agent in state space and action space. The state space of the continuous agent includes the residual data size matrix of RSUs at $t^\text{th}$ time slot $\mathbf{U}_\text{res}[t]$, the channel gain matrix $\mathbf{h}[t]$ and current UAV location $(x_\text{uav}[t], y_\text{uav}[t])$, while the state space of the discrete agent only includes $\mathbf{U}_\text{res}[t]$ and $\mathbf{h}[t]$ because the UAV location is not directly related to the channel allocation.
The output actions of the two agents are also different. The output action of the discrete agent is a set of probabilities $\mathbf{Pr}[t]$ that correspond to the discrete actions, and the decision is made by sampling from the probability set. The output action of the continuous agent on the $x$-axis and $y$-axis is the mean value ($\mu_x,\mu_y$) and variance ($\sigma_x^2,\sigma_y^2$) of two Gaussian distributions $N(\mu_x,\sigma_x^2)$ and $N(\mu_y,\sigma_y^2)$, from which the trajectory action is sampled.

Once trajectory, power control, and channel allocation actions are generated, a combined decision containing these three actions is adopted by the UAV and RSU to interact with the environment. The environment gives rewards to both the continuous agent and discrete agent for the next training step according to the received action.

\subsection{Simulation Results}
Consider a UAV-assisted data collection network with three orthogonal communication channels and maximum transmission power $P_\text{max}=5\text{W}$. The data size to be collected at each RSU is set to $50\text{Mb}$. The proposed framework adopts proximal policy optimization (PPO) algorithm for both the discrete agent and the continuous agent and is tested against three benchmark algorithms: (1) DQN-PPO algorithm adopts deep $Q$-learning (DQN) algorithm \cite{DQN} for the discrete agent and PPO algorithm for the continuous agent; (2) Duelling DQN-PPO algorithm adopts Duelling DQN algorithm for the discrete agent and PPO algorithm for the continuous agent; (3) PPO-A2C algorithm adopts PPO algorithm for the discrete agent and advantage actor-critic (A2C) algorithm for the continuous agent.

\begin{figure}[htb]
\centering
\includegraphics[width=0.9\linewidth]{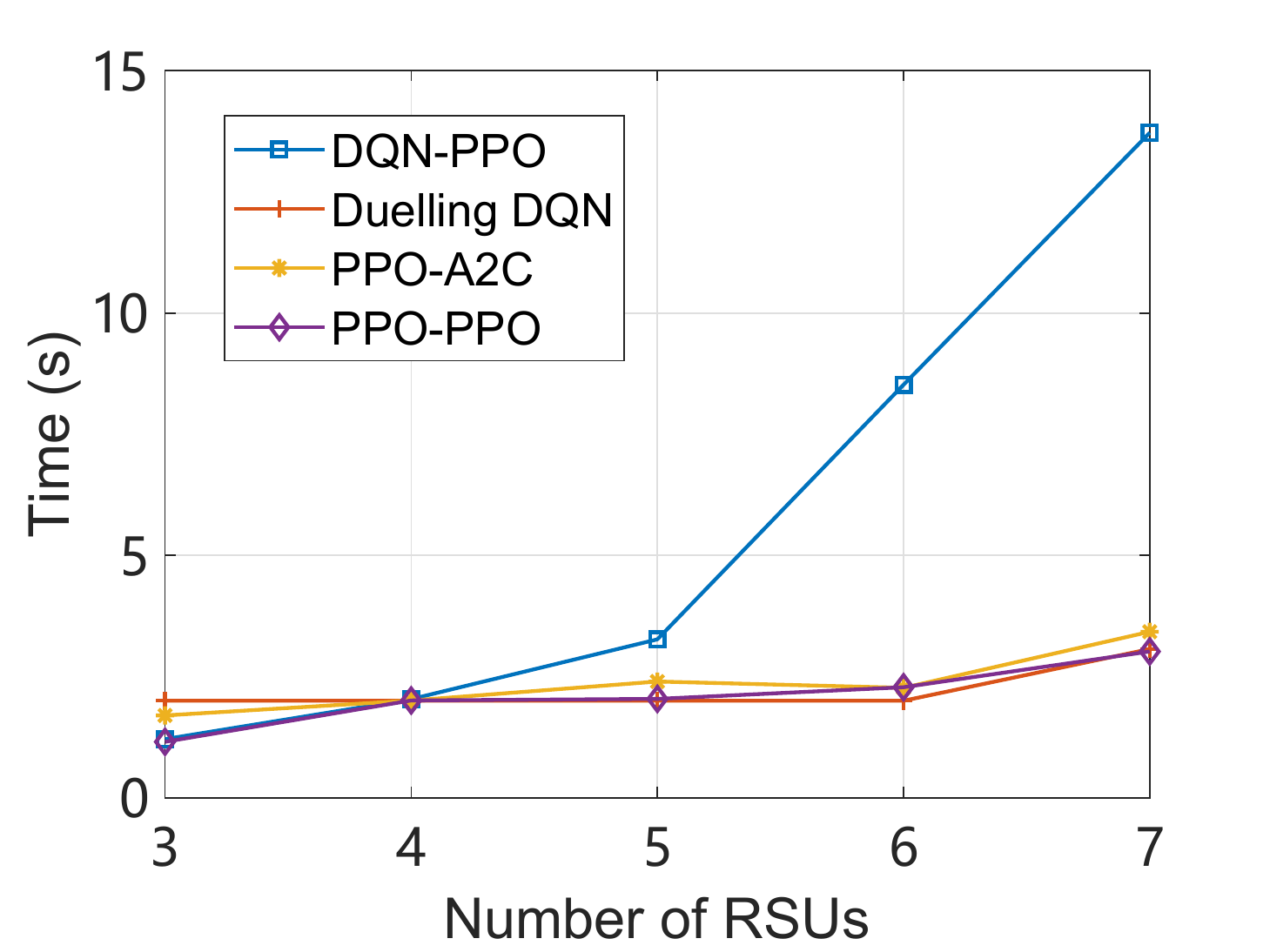}
\caption{Required time to collect data from different numbers of RSUs.}
\label{fig:simu_time}
\end{figure}

Fig. \ref{fig:simu_time} presents the average required time and reward in the last 1000 episodes of the training process with different algorithms and numbers of RSUs (denoted by $N$). We can find a positive correlation between required mission time and the number of RSUs with all of the four algorithms in the figure.  The increase of $N$ results in an increase of total data size, which takes more time to be transmitted. The DQN-PPO algorithm shows the worst performance when the number of RSUs increases due to its poor generalization ability to complex optimization problems. There is a slight difference in the required mission time of the rest of the algorithms. The PPO-PPO algorithm requires less mission time than PPO-A2C and Duelling DQN algorithms when $N=3$, and also shows good performance when $N$ increases.
Thus, the PPO-PPO algorithm is considered more robust against the change in the number of RSUs than the benchmark algorithms.

\begin{figure}[htb]
\centering
\includegraphics[width=0.9\linewidth]{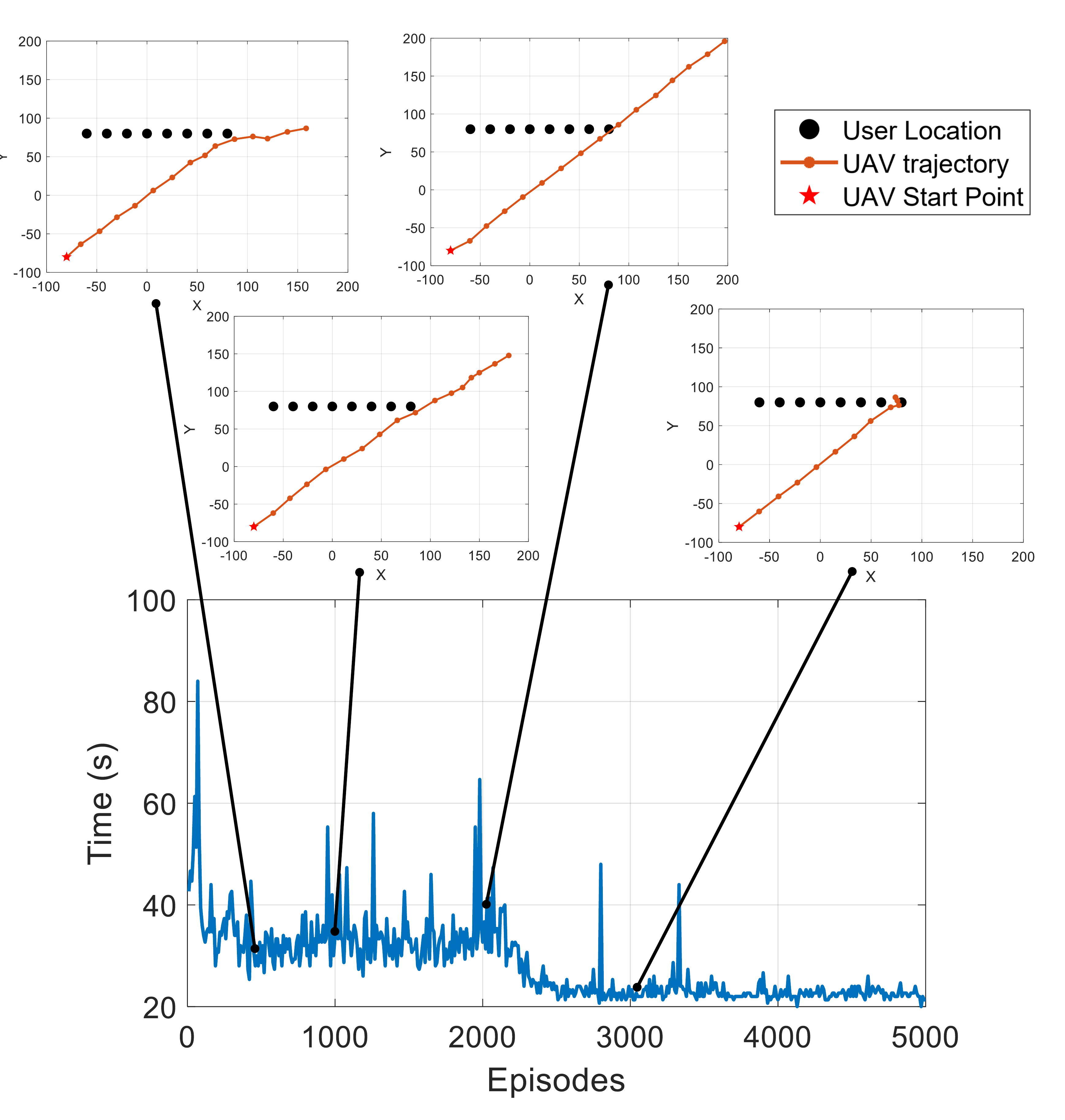}
\caption{An example of UAV trajectory with manually set RSU locations.}
\label{fig:simu_traj}
\end{figure}

The change in UAV trajectory during the training process is shown in Fig. \ref{fig:simu_traj}. In this experiment, we consider the case where RSUs are deployed along a straight road, and locate at $(-60, 80)$, $(-40, 80)$, $(-20, 80)$, $(0, 80)$, $(20, 80)$, $(40, 80)$, $(60, 80)$ and $(80, 80)$. The policy of agents changes over the training process, and we sample the UAV trajectory at different stages ($500^\text{th}$, $1000^\text{th}$, $2000^\text{th}$ and $3000^\text{th}$ episodes).
At the beginning of training ($500^\text{th}$ episode), the UAV adopts a sub-optimal policy, i.e., to fly past the rightmost MDC and keep going further. As the training process goes on, the agent keeps exploring and learning the best policy, and there is a  gradual change in the UAV trajectory: it becomes more straight, and in the final period of the training process ($3000^\text{th}$ episode) the UAV hovers near the rightmost MDC for better channel condition, which results in faster data collection and shorter trajectory. Corresponding to the change in UAV trajectory, the required mission time decreases gradually and finally converges at a small value.

\section{Current Challenges and Future Research Directions}

\begin{figure*}[htb]
 \centering
 \includegraphics[width=0.9\linewidth]{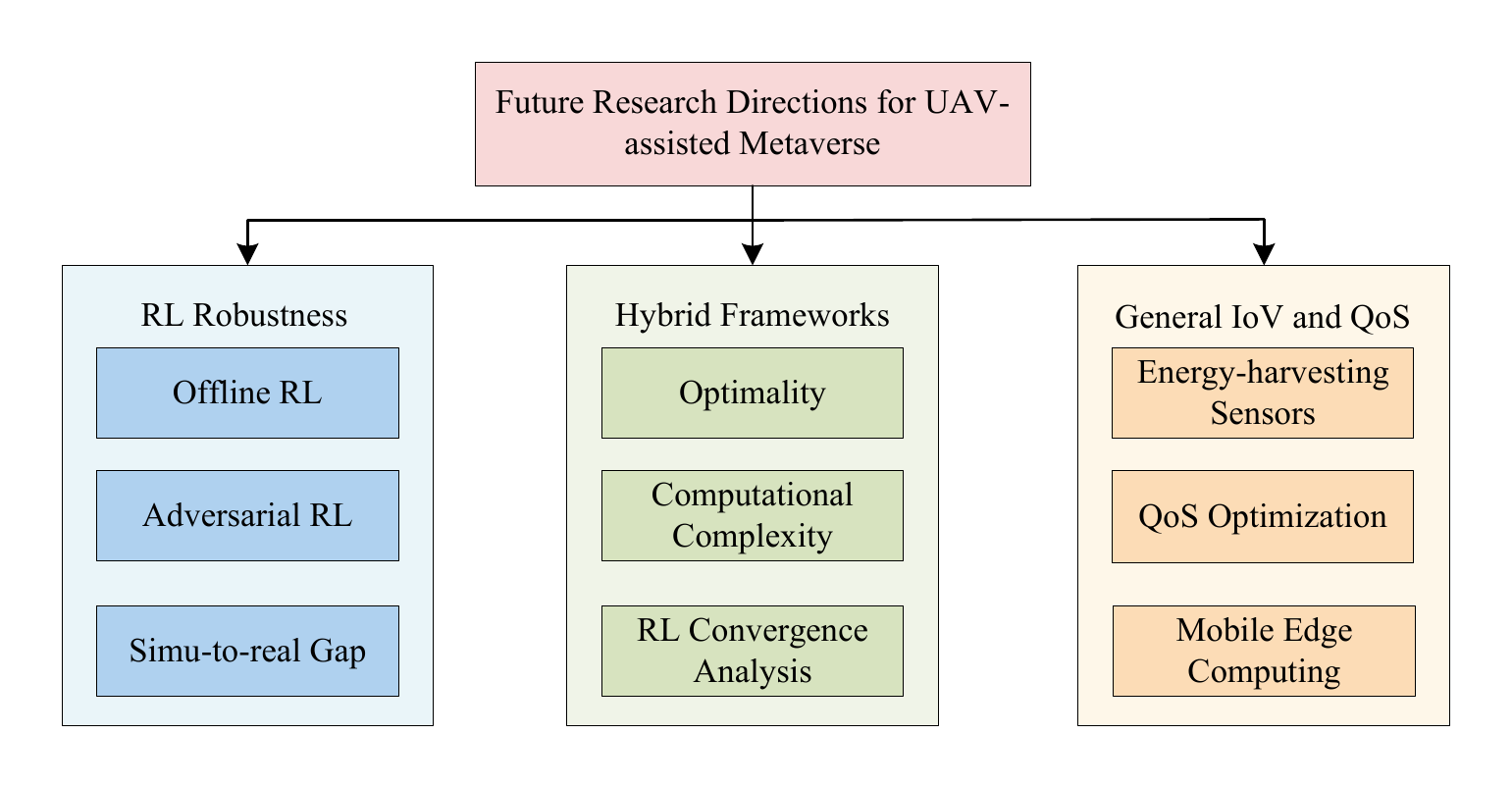}
 \caption{Future Research Directions for UAV-assisted Metaverse.}
 \label{fig:future}
\end{figure*}

The proposed hybrid framework with reinforcement learning and convex optimization shows its potential in solving communication optimization problems, but there are still challenges and future directions to be explored.

%robustness of RL, security adversarial
\subsection{Robustness of Reinforcement Learning}
In recent years, reinforcement learning algorithms and frameworks have been developed rapidly. New algorithms, e.g., DQN, Duelling DQN, A2C, and PPO, that are more robust and powerful have been derived from the original $Q$-learning algorithm. Upon the proposal of each algorithm, extensive experiments in classical test environments will be conducted to verify its performance improvement.
% problem
However, some research indicates that the success of simulation in test environments is not sufficient to ensure the robustness of applications in the physical world, and there is still a large space for improvement of the robustness and security of reinforcement learning algorithms\cite{robust_RL_PMLR}.
\subsubsection{Simulation-to-reality (Simu-to-real) Gap}
    Training the reinforcement learning agent by interacting with real environments is costly because the policy is not stable at the beginning of the training process. Thus, the training process of reinforcement learning agents is usually based on ideal simulation environments, i.e., prior assumptions on data distribution and state transition are given, which leads to the distribution gap between test sets and real data sets. Similar to the overfitting phenomenon in traditional machine learning, this \emph{simu-to-real gap} could lead to a huge drop of performance in physical world applications. Further research on more powerful algorithms is required to improve the robustness of the existing reinforcement learning algorithms.

\subsubsection{Adversarial Reinforcement Learning}
    The action decision made by reinforcement learning agents is sensitive to observation, which makes it vulnerable to adversarial attacks. Attacks on the observation of agents work in a way similar to generative adversarial networks (GAN), which aim to mislead the agent to make worse decisions by adversarial observations while reducing the possibility of being detected. Another type of attack is to modify the output action of the agent, which could be affected by malicious controllers or uncertain disturbance caused by the environment. To obtain a robust policy under such attacks, it is of vital importance to estimate the uncertainty of the environment and minimize its effect on action decisions by robust algorithms such as Probabilistic Action Robust Markov Decision Process (PR-MDP) and Noisy Action Robust MDP (NR-MDP).
\subsubsection{Offline Reinforcement Learning}
Most existing reinforcement learning (RL) algorithms are based on online learning, i.e., training the agent while interacting with the real environment. This paradigm enables efficient data collection and neural network training, but it is also an obstacle to the widespread application in fields where mistakes are costly and dangerous. It is critical to find a training method for RL agents without interacting with the environment with the sub-optimal and unstable policy that is not fully trained.

The success of traditional machine learning motivated researchers to develop offline reinforcement learning, which trains the agent with prepared data/records of interaction with the environment, and the training process is free from interaction with the real environment.
Offline reinforcement learning is a promising alternative to traditional online reinforcement learning to avoid possible accidents during the training process, but it still faces critical algorithmic challenges. Firstly, learning from the prepared offline data is not sufficient to learn about the entire environment information, and there could be deviations between the training data set and the real data set. Secondly, the out-of-distribution (OOD) data in the training data set can be fatal to the policy training, which results in the collapse of algorithms. There is still a long way to reliable and efficient offline reinforcement learning algorithm.

\subsection{Trade-off between Reinforcement Learning and Convex Optimization}
%complexity estimation
The combination of both convex optimization and reinforcement learning brings the trade-off problem in framework
designing.
Convex optimization algorithms are mature and reliable methods with a strict mathematical proof of the optimal solution, but their implementation and optimality proving can be hard in complex time-sequential problems. Reinforcement learning is more capable to handle complex problems, but it faces problems of robustness and generalization. How to make the perfect combination of them to solve complex problems is a topic that still needs to be studied and discussed.
In this section, some metrics are given as references for framework designing.

\subsubsection{Optimality Concerns}
    As for the problems that can be solved by convex optimization algorithms to obtain the global optimal, traditional convex optimization methods are still the first option. But those who are complex and time-sequential, reinforcement learning should be taken into consideration for possible solutions.
\subsubsection{Computational Complexity}
    The required computing time should be considered when designing a hybrid framework with reinforcement learning and convex optimization because it directly affects the latency of decisions.
    The complexity of traditional convex optimization algorithms can be evaluated and formulated by frameworks such as oracle complexity \cite{oracle_complexity}. For example, for an unconstrained optimization problem
    $\underset{\mathbf{w}\in {{\mathbb{R}}^{d}}}{\mathop{\min }}\,f(\mathbf{w})$,
where $f$ is a generic smooth and convex function and has $
\mu_1$-Lipschitz gradients and the algorithm makes its first oracle query at a point $\mathbf{w}_1$, the complexity to obtain a solution $\mathbf{w}_T$ that satisfies $f({{\mathbf{w}}_{T}})-{{\min }_{\mathbf{w}}}f(w)\le \varepsilon$ is given by
${\mathrm O}\left( \sqrt{({{\mu }_{1}}{{D}^{2}})/\varepsilon } \right)$,
where $D$ denotes the upper bound of the distance between $\mathbf{w}_1$ and the nearest minimizer of $f$ \cite{oracle_complexity}.
The complexities of ``model-based" reinforcement learning algorithms are usually evaluated by ``sample complexity". ``Model-based" means that the transitions in the MDP is known by the agent, and ``sample complexity" is defined as the required high probability upper bound on the number of training steps to make the algorithm not nearly optimal. Most state-of-art reinforcement learning algorithms are ``model-free", i.e., the environment's information is hidden. Deep reinforcement learning (DRL) algorithms are widely applied in unknown environments, and their complexity can be evaluated by the required computation in a single step, including the action generation process and the training process. Consider a DRL-based agent with an input layer of size $Z_0$ and $L$ training layers with sizes $Z_1, \ldots, Z_L$, the complexity in a single time step is ${\mathrm O}\left( {{Z}_{0}}{{Z}_{1}}+\sum\nolimits_{l=1}^{L-1}{{{Z}_{l}}{{Z}_{l+1}}} \right)$.

\subsubsection{Convergence of Reinforcement Learning}
The convergence of reinforcement learning algorithms have direct influence on the final performance of the hybrid framework. Currently, the study on convergence has started in safe RL, which is defined as maximizing the expected reward while satisfying given constraints. Safe RL algorithms can be categorized into primal and primal-dual methods. There has been convergence analyses on primal-dual safe RL, but few works have been done to explain the convergence of primal safe RL. Further study of the convergence issue on RL is still in demand for reference of building hybrid frameworks with RL and convex optimization methods.
% convex-RL current study LSTM

\subsection{General IoV and QoS Optimization}
The case study in this paper discusses a specific case, but further study on UAV-assisted Metaverse can be extended to general IoV and QoS optimization, including energy-harvesting wireless sensors, mobile edge computing, and other types of vehicles.
\subsubsection{Energy-harvesting Wireless Sensors}
The sensors deployed at RSUs in the case study of this paper are assumed to be supported by sufficient energy, {but in some cases, the sensors are designed to be self-charging, i.e., harvesting energy from the environment and storing it in local batteries. In this context, there are promising technologies to tackle this issue such as wake-and-sleep policy and ambient backscatter communications, where the backscatter tag could harvest signals from surrounding environments}, e.g., RSUs, base stations, and WiFi access points, to transmit messages to the receiver with any dedicated radio frequency sources. With this novel idea, less-power-consumption or battery-less communication schemes would be achieved. Besides this, reconfigurable intelligent surface (RIS) is considered another potential approach to capturing energy efficiency through passive RIS units. In this model, instant signals reflected from RISs could be adaptively enhanced by programmatically modifying their phase shifts with no additional power. For those reasons, these two technologies potentially contribute to the future metaverse networks. However, as the limitation of batteries and the energy-harvesting rate is not considered in this paper, further study could be conducted to explore the energy-efficient methods to complete the given data collection mission.
\subsubsection{QoS Optimization}
The Metaverse aims to provide service for massive users with heterogenous demand and evaluation standards for QoS, including transmission delay, throughput, energy efficiency, and data collection mission completion time. In this case, the single-standard-based policy can not efficiently satisfy the demand of different users and obtain the best decision to maximize the QoS. The virtual service provider (VSP) should be able to switch between various objective functions, and hybrid-standard for QoS evaluation should be adopted by the reward setting of reinforcement learning agents to balance the cost and income.
\subsubsection{Mobile Edge Computing}
The Metaverse is expected to implement seamless immersive experience with the support of VR/AR devices. An obstacle in the way to Metaverse is the limited computation capacity of mobile devices, which motivates the study of mobile edge computing (MEC). MEC aims to fully utilize the computational capacity of edge IoT devices, and to find the balance among calculation delay, communication latency, and power consumption for better general performance and user experience.  Although there have been works on MEC in traditional applications, its usage in Metaverse is rarely touched. Thus, further study on MEC is required to pave the way for Metaverse.
\section{Conclusion}
The development of reinforcement learning brings new methods to solve complex optimization problems that are difficult to solve by traditional convex optimization algorithms. In this paper, we propose a hybrid framework with reinforcement learning and convex optimization, which decomposes the original problem into multiple sub-problems and solves them by reinforcement learning agents and convex solvers cooperatively. A case study based on UAV-assisted data collection in Metaverse networks shows the feasibility of solving the joint resource allocation and UAV trajectory optimization problems with our proposed framework. \textcolor{black}{Current challenges and future research directions for the case study have been elaborated on.}


\begin{thebibliography}{99}
\bibitem{cheng2022will}
R. Cheng, N. Wu, S. Chen, and B. Han, ``Will metaverse be nextG Internet vision, hype, and reality," \emph{IEEE Network}, vol. 36, no. 5, pp. 197-204, 2022.
\bibitem{Vision_UAV_meta}
M. Xu, W. C. Ng, W. Y. B. Lim, J. Kang, Z. Xiong, D. Niyato, Q. Yang, X. S. Shen, and C. Miao, ``A full dive into realizing the edgeenabled metaverse: Visions, enabling technologies, and challenges,"
\emph{IEEE Communications Surveys \& Tutorials}, 2022.
\bibitem{uav_base_station}
S. Wan, J. Lu, P. Fan, and K. B. Letaief, ``Toward big data processing in IoT: Path planning and resource management of UAV base stations in mobile-edge computing system," \emph{IEEE Internet of Things Journal}, vol. 7, no. 7, pp. 5995-6009, 2019.
\bibitem{uav_data_collect_survey}
Z. Wei, M. Zhu, N. Zhang, L. Wang, Y. Zou, Z. Meng, H. Wu, and Z. Feng, ``UAV assisted data collection for internet of things: A survey,"
\emph{IEEE Internet of Things Journal}, 2022.
\bibitem{metaverse_uav}
Y. Han, D. Niyato, C. Leung, C. Miao, and D. I. Kim, ``A dynamic resource allocation framework for synchronizing metaverse with IoT service and data," \emph{ICC 2022-IEEE International Conference on Communications}, pp. 1196-1201, 2022.
\bibitem{energy_aware_UAV}
L. Shen, N. Wang, D. Zhang, J. Chen, X. Mu, and K. M. Wong, ``Energy-aware
dynamic trajectory planning for UAV-enabled data collection in
mMTC networks," \emph{IEEE Transactions on Green Communications and
Networking}, vol. 6, no. 4, pp. 1957-1971, 2022
\bibitem{time_limit_UAV}
K. Liu and J. Zheng, ``UAV trajectory optimization for time-constrained
data collection in UAV-enabled environmental monitoring systems,"
\emph{IEEE Internet of Things Journal}, vol. 9, no. 23, pp. 24 300-24 314,
2022.
\bibitem{MINLP_survey}
S. Burer and A. N. Letchford, ``Non-convex mixed-integer nonlinear programming:
A survey," \emph{Surveys in Operations Research and Management
Science}, vol. 17, no. 2, pp. 97-106, 2012.
\bibitem{rl_v2v}
H. Ye, G. Y. Li, and B.-H. F. Juang, ``Deep reinforcement learning based
resource allocation for V2V communications," \emph{IEEE Transactions on
Vehicular Technology}, vol. 68, no. 4, pp. 3163-3173, 2019.
\bibitem{RL_multi}
J. Cui, Y. Liu, and A. Nallanathan, ``Multi-agent reinforcement learningbased
resource allocation for UAV networks," \emph{IEEE Transactions on
Wireless Communications}, vol. 19, no. 2, pp. 729-743, 2019.
\bibitem{actor-critic}
I. Grondman, L. Busoniu, G. A. Lopes, and R. Babuska, ``A survey of
actor-critic reinforcement learning: Standard and natural policy gradients,"
\emph{IEEE Transactions on Systems, Man, and Cybernetics, Part C
(Applications and Reviews)}, vol. 42, no. 6, pp. 1291-1307, 2012.
\bibitem{PPO}
J. Schulman, F. Wolski, P. Dhariwal, A. Radford, and O. Klimov, ``Proximal
policy optimization algorithms," \emph{arXiv preprint arXiv:1707.06347},
2017.
\bibitem{DQN}
J. Fan, Z. Wang, Y. Xie, and Z. Yang, ``A theoretical analysis of deep
Q-learning," \emph{in Learning for Dynamics and Control. PMLR}, 2020, pp.
486-489.
\bibitem{robust_RL_PMLR}
L. Pinto, J. Davidson, R. Sukthankar, and A. Gupta, ``Robust adversarial
reinforcement learning," \emph{in International Conference on Machine
Learning. PMLR}, 2017, pp. 2817-2826.
\bibitem{oracle_complexity}
Y. Arjevani, O. Shamir, and R. Shiff, ``Oracle complexity of second-order
methods for smooth convex optimization," \emph{Mathematical Programming},
vol. 178, no. 1, pp. 327-360, 2019.

\end{thebibliography}
\end{document}